\documentclass{article}

\usepackage{PRIMEarxiv}
\usepackage{amsmath,amsfonts}
\usepackage{algorithmic}
\usepackage{algorithm}
\usepackage{array}
\usepackage[caption=false,font=normalsize,labelfont=sf,textfont=sf]{subfig}
\usepackage{textcomp}
\usepackage{stfloats}
\usepackage{url}
\usepackage{verbatim}
\usepackage{graphicx}
\usepackage[numbers]{natbib}

\usepackage{tabularx}
\usepackage{siunitx}
\usepackage{hyperref}       
\usepackage{url}    
\usepackage{booktabs} 
\usepackage{tabularx}
\usepackage{makecell}
\usepackage{multirow}
\title{TreeLearn: A deep learning method for segmenting individual trees from ground-based LiDAR forest point clouds}

\newcolumntype{Y}{>{\centering\arraybackslash}X}
\newcolumntype{Z}{>{\scriptsize}X}
\newcommand{\errom}{E_{\text{om}}}
\newcommand{\errcom}{E_{\text{com}}}
\newcommand{\argmax}{\operatornamewithlimits{argmax}}
\newcommand{\fig}[1]{Figure~\ref{#1}}
\newcommand{\secref}[1]{Section~\ref{#1}}
\newcommand{\tableref}[1]{Table~\ref{#1}}

\usepackage{authblk}
\author[1, *]{Jonathan Henrich}
\author[2, *]{Jan van Delden}
\author[3]{Dominik Seidel}
\author[1]{Thomas Kneib}
\author[2, 4]{Alexander S. Ecker}
\affil[1]{Chairs of Statistics and Econometrics,
Faculty of Economics,
  University of Göttingen}
\affil[2]{ Institute of Computer Science,
  University of Göttingen}
\affil[3]{Department Spatial Structures and Digitization of Forests,
  Faculty of Forest Science,
  University of Göttingen}
\affil[4]{Campus Institute Data Science, University of Göttingen and
  Max Planck Institute for Dynamics and Self-Organization,
  Göttingen}

\begin{document}

\maketitle

\begin{abstract}
Laser-scanned point clouds of forests make it possible to extract valuable information for forest management. To consider single trees, a forest point cloud needs to be segmented into individual tree point clouds. 
Existing segmentation methods are usually based on hand-crafted algorithms, such as identifying trunks and growing trees from them, and face difficulties in dense forests with overlapping tree crowns. In this study, we propose TreeLearn, a deep learning-based approach for tree instance segmentation of forest point clouds. TreeLearn is trained on already segmented point clouds in a data-driven manner, making it less reliant on predefined features and algorithms. Furthermore, TreeLearn is implemented as a fully automatic pipeline and does not rely on extensive hyperparameter tuning, which makes it easy to use.
Additionally, we introduce a new manually segmented benchmark forest dataset containing 156 full trees. The data is generated by mobile laser scanning and contributes to create a larger and more diverse data basis for model development and fine-grained instance segmentation evaluation. We trained TreeLearn on forest point clouds of \num[group-digits=integer]{6665} trees, labeled using the Lidar360 software. An evaluation on the benchmark dataset shows that TreeLearn performs as well as the algorithm used to generate its training data. Furthermore, the performance can be vastly improved by fine-tuning the model using manually annotated datasets. We evaluate TreeLearn on our benchmark dataset and the Wytham Woods dataset, outperforming the recent SegmentAnyTree, ForAINet and TLS2Trees methods. The TreeLearn code and all datasets that were created in the course of this work are made publicly available.
\end{abstract}

{
\renewcommand{\thefootnote}{*} 
\footnotetext{The first two authors contributed equally to this work.
\newline Correspondence: \texttt{jonathan.henrich@uni-goettingen.de}}
}

\section{Introduction}

\begin{figure}[t]
\includegraphics[width=\textwidth]{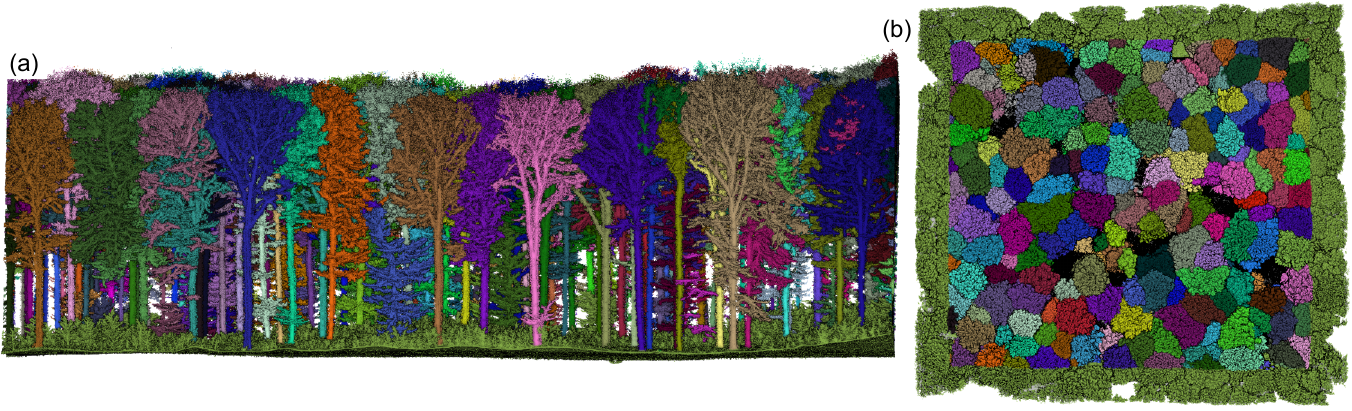}
    \caption[]{(a): predictions generated by TreeLearn on the benchmark forest dataset. (b): a top-view of the manually segmented benchmark forest dataset. For the green area at the edge there are no segmentation labels. It is still included in the dataset since it provides relevant context information for the labeled segment.}
    \label{fig:intro}
\end{figure}

Today, forests are not only exposed to an ever-growing pressure from human use but also to a rapidly changing environment due to climate change. To understand changes in forest composition and structure as a response to these stressors, as well as for successful forest management, scientists and forest managers require precise and detailed information on the current status of our forests. To assess the structural characteristics of forests or individual trees, laser scanning techniques are a powerful tool \cite{liang2014use, disney2019terrestrial, calders2020terrestrial, yun2024status}. They allow to create high-resolution 3D point clouds of the surfaces in a forest. To analyze such data, it is useful to separate the point cloud into the individual trees (\fig{fig:intro}a), e.g. to classify them by their species \cite{terryn2020tree, xi2020see, seidel2021predicting} or to estimate the above-ground biomass \cite{xu2021lidar, brede2022non}. Separating a point cloud into individual trees is an instance segmentation problem: tree points must first be identified in the point cloud and then each point must be assigned to an individual tree instance. Since manually segmenting forest point clouds is highly time-consuming and subjective, there is a need for methods that automate this process. The method proposed in this work tackles this task for point clouds acquired using laser scanners that are able to capture forest structure below the canopy. Primary sources of such point clouds are terrestrial and mobile laser scanning (TLS and MLS), but also laser scanning via low-flying unmanned aerial vehicles (UAV). Point clouds obtained via airborne laser scanning (ALS) do not capture the lower parts of the forest in high detail and therefore require a different set of methods, e.g. \cite{lei2022novel, cao2023benchmarking, fu2024individual}.

The automatic segmentation of MLS and TLS point clouds into individual tree instances is a relatively young research field that has so far been mostly tackled by algorithms based on hand-crafted features. These algorithms usually first identify tree trunks and then assign the remaining points to these trunks based on a fixed set of rules \cite{tockner2022automatic, deng2024individual}. In early works, these rules were relatively simple. For example, some authors divided the point cloud into clusters and merged them based on distance and relative orientation~\cite{trochta20173d,raumonen2015massive}. Follow-up work building upon these methods included more elaborate local geometry and shape features~\cite{burt2019extracting}. More recently, \citet{fu2022segmenting} avoided complicated feature computations and proposed to cluster trees from bottom to top in a layer-by-layer manner. Another strand of literature conceptualized the forest as a graph and segmented trees via graph cut \cite{zhong2016segmentation, heinzel2018constrained, xi20223d} or graph pathing methods \cite{wang2021individual}. Furthermore, some works aimed to incorporate biological theories. For example, \citet{liu2021point} constructed features based on plant morphology theory which were then used in a region-growing approach. Similarly, \citet{tao2015segmenting} developed a shortest-path algorithm based on metabolic theory, which states that plants tend to minimize the transferring distance to the root. \citet{wang2020unsupervised} also employed metabolic theory by constructing a superpoint graph where the node features depend on shortest path analysis. Recently, \citet{xu2023topology} proposed a segmentation method that identifies distinctive features in the forest and extracts individual trees based on topological structure. Compared to previous approaches, the method is more convenient for practitioners as it reduces the need for intensive hyperparameter tuning.

Although modern frameworks achieved significant improvements in performance, automated tree instance segmentation remains prone to errors \cite{martin2021evaluation}, especially in dense forests with heavily intersecting tree crowns. In such scenarios, hand-crafted features and pre-defined heuristics appear to be too inflexible to account for the plethora of possible interactions between trees. Machine learning approaches could overcome these limitations, since features and association rules are derived directly from the data through gradient-based learning. In fact, recent advances in point cloud processing have been dominated by machine learning methods, as indicated by their performance on various point cloud processing benchmarks, e.g. \cite{armeni_cvpr16, dai2017scannet, hackel2017semantic3d}.

Although machine learning methods have been applied to the segmentation of trees from forest point clouds ~\cite{wang2019individual, windrim2019forest, windrim2020detection,chang2022two}, they mostly rely on projections into 2D. As a result, they lose information, because the network does not operate on the original point cloud. Other works do employ 3D deep learning for semantic segmentation, but resort to classical algorithms for tree instance segmentation \cite{chen2021individual, krisanski2021forest, wilkes2022tls2trees, wielgosz2023point2tree}. To our knowledge, there are only a few studies in the related field of urban tree segmentation that actually employ 3D deep learning methods for tree instance segmentation \cite{luo2021individual, jiang2023segmentation}. Furthermore, the concurrent works ForAINet \cite{xiang2024automated} and SegmentAnyTree \cite{wielgosz2024segmentanytree} recently showed the potential of directly performing deep-learning-based instance segmentation on forest point clouds. It is surprising that this research direction has not been explored more thoroughly since state-of-the-art instance segmentation methods for 3D point clouds exclusively operate on 3D input data~\cite{vu2022softgroup, jiang2020pointgroup, sun2022superpoint, schult2022mask3d, wu20223d, ngo2023isbnet}, making them promising candidates for improving existing tree segmentation methods. In this paper, we adapt one of these methods~\cite{jiang2020pointgroup} and propose TreeLearn, a deep learning method to segment trees from forest point clouds. Our method first identifies tree points in the point cloud, then projects them towards the respective tree base, and groups them via density-based clustering. The density-based clustering is performed at once on the whole input such that complete tree instances are obtained in a single step. This is a crucial difference to the concurrent works ForAINet and SegmentAnyTree where tree instances are obtained by merging multiple incomplete tree predictions, which can lead to errors such as fragmentation. Our method is realized in an automatic pipeline which can be used out-of-the-box without extensive hyperparameter tuning. This reduces the number of work steps required to be performed by the user to a minimum, which makes our method easy to apply. Our method achieves state-of-the-art results on two tree instance segmentation datasets, outperforming the recent similar models ForAINet and SegmentAnyTree.

Since data-driven deep learning methods require large amounts of data for parameter optimization, high-quality labeled forest point clouds are urgently needed. Although there are some studies that use forest point clouds with manually corrected segmentation labels, some of them do not make their datasets public \cite{heinzel2018constrained, xi20223d}. Furthermore, those studies that provide publicly available datasets \cite{calders2014terrestrial, weisertreedataset, tockner2022automatic, cao2023benchmarking, puliti2023instance} often have properties which limit their use for training machine learning methods. For example, \citet{calders2014terrestrial} only provides segmentation labels for a small number of trees located within a larger forest plot without segmentation labels. \citet{weisertreedataset} acknowledge quality limitations of their tree segmentations as well as low laser scanning coverage. Other studies only provide labeled trees instead of the complete point clouds \cite{tockner2022automatic, calders2022laser, cao2023benchmarking}. We argue that machine learning methods should be able to handle the complete registered point clouds to make its usage as easy as possible for practitioners. Furthermore, the separation of tree points from understory and ground points is often non-trivial and therefore itself an interesting training objective. A concurrent work \cite{puliti2023instance} acknowledges the need for complete labeled forest point clouds by providing a large-scale dataset with high-quality semantic and instance segmentations of 1130 trees acquired from UAV scans. In our work, we also provide a benchmark dataset of a continuous forest with high-quality manual crown segmentations and classification of points into non-tree and tree. It contains 156 trees that are completely within the labeled forest area (\fig{fig:intro}b). Furthermore, we extend the publicly available labeled trees in the LAUTx \cite{tockner2022automatic} and Wytham Woods \cite{calders2022laser} datasets by refining the segmentation labels and propagating them to the complete point cloud. These dataset contributions are an important step towards creating a rich data basis for the development of powerful and general machine-learning-based tree segmentation algorithms. Additionally, the datasets can be used to systematically evaluate and compare existing methods.

In summary, our study has three main contributions:

\begin{itemize}
\item TreeLearn, a fully automatic deep-learning-based pipeline that takes forest point clouds as input, identifies tree points, and groups them into individual trees.
\item A high-quality hand-segmented forest dataset that can be used to train and systematically evaluate machine learning methods for tree segmentation.
\item A thorough comparison between TreeLearn and other models (Lidar360, TLS2Trees, ForAINet and Segment\-AnyTree) on our benchmark dataset and the Wytham Woods dataset.
\end{itemize}

\section{Material and methods}

In the following, we describe both the datasets used in this study and our TreeLearn approach in detail. Readers unfamiliar with the technical background can skip directly to the results section.

\subsection{Datasets}

The data basis for model training, validation and evaluation consists of two parts: (1) We provide MLS point clouds from 19 forest plots located in Germany (\textit{our data}). For one of these plots, we performed manual segmentation, ensuring a clean and controlled separation of crowns. The other plots were automatically segmented using the Lidar360 software \cite{Lidar360}. (2) The publically available LAUTx \cite{tockner2022automatic}, Wytham Woods \cite{calders2022laser}, and FOR-Instance \cite{puliti2023instance} datasets (\textit{additional data}). The LAUTx and Wytham Woods datasets do not originally contain the understory and ground component but only individual segmented trees, complicating their application for training and evaluation of our pipeline. We refined and propagated the tree labels of these datasets to the full point clouds which we obtained by contacting the authors. In Sections \ref{section:ourdata} and \ref{section:additional_data}, we describe both our data and the additional data. Section~\ref{sec:training} explains how these datasets were used for training, validation, and testing. All datasets that were created or extended in the course of this work are made publicly available as LAS files. LAS files allow to save point cloud coordinates together with labels. Following \citet{puliti2023instance}, we use the \textit{treeID} and \textit{classification} field to distinguish tree points, non-tree points and non-annotated points.

\subsubsection{Our data}\label{section:ourdata}

Our data consists of 19 MLS point clouds located in Germany. Eight plots are located near the city of Göttingen, in Lower Saxony, two near Allstedt in Saxony-Anhalt, two near Oppershofen in Hesse, and seven near Lübeck in Schleswig-Holstein. The stands are all dominated by European beech (Fagus sylvatica L.) and between 92 and 162 years of age.  The size of the study plots ranges from 1.0 to 2.2~ha. {Ten plots have been managed for decades, for the others management has been abandoned for at least 30 years. More detailed information on the individual forest plots can be found in Appendix~\ref{app:dataset}.

\paragraph{Data acquisition}

The plots were recorded in February 2021, in leaf-off condition, using a ZEB Horizon mobile laser scanner (Geoslam Ltd., Nottingham, UK). The device uses a laser to measure the distance and direction to surrounding objects while being carried through a scene. Based on the principle of simultaneous localization and mapping (SLAM) the ZEB Horizon creates a 3D point cloud of the scenery up to a distance of 100~m from the walking trajectory of the operator. Each hand-held walkthrough with the scanner resulted in a file that was used to create a point cloud by conducting the spatial coregistration (actual SLAM) in GeoSlam HUB Version 6 \cite{GeoSLAM_HUB_V6}. The resulting point clouds were subsampled to a resolution of 1~cm to match the effective resolution of the ZEB Horizon mobile laser scanner. Additionally, the statistical outlier removal functionality provided by CloudCompare \cite{cloudcompare} was used to remove points that are not of interest, such as scanned particles in the air.

\paragraph{Segmentation labels} \label{sec:Lidar360labels}

We obtained our segmentation labels for the forest point clouds in two ways: (1) by using the automatic segmentation functionality provided by the Lidar360 software \cite{Lidar360} and (2) by manually correcting the segmentation labels of a single forest plot. 

\paragraph{Automatic segmentation}
To automatically segment the 19 forest point clouds into individual trees, we used the TLS-package from Lidar360 as described by \citet{neudam2023simulation}. Lidar360 employs an implementation of the comparative shortest-path algorithm from \citet{tao2015segmenting} for segmenting the point clouds. The hyperparameters of the Lidar360 software were optimized to produce visually appealing tree segmentations. A minimum size of 10~m was chosen as a requirement for tree classification. Consequently, vegetation that is smaller than 10~m or points belonging to the forest ground were classified as non-tree points. Lidar360's segmentation algorithm operates on a point cloud that is terrain-normalized, i.\,e. has a flat ground surface. This was achieved in a separate preprocessing step. We matched the terrain-normalized points with the original points based on their x- and y-coordinates to obtain labels for the unnormalized point clouds which were used in this study. The automatic segmentation contains numerous errors, including multiple trees detected as a single tree, non-tree points identified as part of trees, and inaccuracies in discerning tree boundaries in crowns and branches that are close to each other (see \secref{sec:results} for qualitative segmentation results).
Since these labels contain numerous mistakes, we refer to them as \emph{noisy labels}. 

\paragraph{Manually corrected plot: L1W}

For one forest plot, L1W, we manually corrected the segmentation labels generated by Lidar360 with the help of CloudCompare. To do this, a python script was used to iterate through all trees in the dataset. In each iteration, the central tree was loaded together with all surrounding trees in CloudCompare and all mistakes that were associated with the central tree were corrected. This protocol ensured that mistakes were tackled one tree at a time to guide the focus of the annotators and avoid sensory overwhelm. (2) Two annotators (authors J. H. and J. v. D.) were involved in the procedure. Trees were manually corrected by one annotator and potentially refined by a second annotator. In case of ambiguities, the two annotators engaged in active discussion.
The corrected segmentation covers an area of \mbox{112 m} by \mbox{103 m}. It includes 200 trees, 156 of which are located entirely within the segment and 44 of which are partly within the segment (\fig{fig:intro}b). The partly labeled trees are also included in the dataset because they can potentially be used for model training since they contain the trunks. However, they should not be used for evaluation since parts of the crown are missing. We appended \mbox{11 m} of the unlabeled forest point cloud on the edges of the labeled segment (\fig{fig:intro}b, green area at the edge). This area is included in the dataset since it provides relevant context information for the manually segmented parts. The ratio of non-tree to tree points in the dataset is approximately 2:3. It should be noted that trees smaller than 10~m were assigned to the non-tree class for this dataset. This is in contrast to the other datasets used in this work (see \secref{section:additional_data}) that also contain annotations for smaller trees.

\subsubsection{Additional data: FOR-Instance, Wytham Woods and LAUTx}\label{section:additional_data}

\begin{table*}[t!] 
\centering
\scriptsize
\caption{Summary of the characteristics of several published labeled forest datasets. Information on NIBIO, NIBIO2, CULS, TU\_WIEN, SCION and RMIT was taken from \cite{puliti2023instance}. The top four rows indicate the datasets that were created or extended in the course of this work. \label{tab:plots}}
\begin{tabularx}{\linewidth}{llZYYYYYY}
\toprule
    & Name  & Country & {Reference} & $n$ plots & $n$ trees & \makecell{Annotated\\area ($ha$)} & Forest type & Sensor \\
\midrule
    & L1W & Germany & \citet{neudam2023simulation} & 1 & 200 & 1.16 & temperate deciduous forest & GeoSLAM ZEB-Horizon \\
    & Lidar360 & Germany & \citet{neudam2023simulation} & 18 & 6665 & 25.09 & temperate deciduous forest & GeoSLAM ZEB-Horizon \\
    & LAUTx & Austria & \citet{tockner2022automatic} & 6 & 514 & 0.83 & temperate mixed forest & GeoSLAM ZEB-Horizon\\
    & Wytham & England & \citet{calders2022laser} & 1 & $\text{877}^{\text{a}}$ & 1.52 & temperate deciduous forest & Riegl VZ-400\\
    \midrule
    & NIBIO & Norway & \citet{puliti2023tree} & 20 & 575 & 1.21 & coniferous dominated boreal forest & Riegl miniVUX-1 UAV\\
    & NIBIO2 & Norway & \citet{puliti2023tree} & 50 & 3062 & 2.87 & coniferous dominated boreal forest & Riegl miniVUX-1 UAV\\
    & CULS & Czech Republic & \citet{kuvzelka2020very} & 3 & 47 & 0.33 & coniferous dominated temperate forest & Riegl VUX-1 UAV\\
    & TU\_WIEN & Austria & \citet{wieser2017case} & 1 & 150 & 0.55 & deciduous dominated alluvial forest & Riegl VUX-1 UAV\\
    & SCION & New Zealand & Unpublished & 5 & 135 & 0.33 & non-native pure coniferous temperate forest & Riegl miniVUX-1 UAV\\
    & RMIT & Australia & Unpublished & 1 & 223 & 0.37 & Native dry sclerophyll eucalypt forest & Riegl miniVUX-1 UAV\\
\bottomrule
\end{tabularx}
\begin{flushleft}
$^{a}$ The trunks of trees in Wytham Woods sometimes diverge directly at the ground. In the original publication, these parts of a tree are labeled with the same number and\\ \phantom{0} a different character (e.g. 1a, 1b, 1c, ...). Unlike other works \cite{wielgosz2024segmentanytree}, we treat these cases as separate instances so that we arrive at 877 trees.
\end{flushleft}
\end{table*}

In addition to our datasets, the existing literature was searched for further forest point clouds with tree segmentation labels. First, there is the recently published FOR-Instance dataset \citep{puliti2023instance} in which tree labels and fine-grained semantic labels were manually added to point clouds from existing works. These point clouds have been captured via UAV-laser scanning and consist of diverse forest plots located in Norway (NIBIO), Czech Republic (CULS), Austria (TU\_WIEN), New Zealand (SCION) and Australia (RMIT). A summary of the characteristics of each dataset can be found in Table~\ref{tab:plots}. More precise information can be found in the respective publications.

Apart from these point clouds, two published datasets were identified that consist of high-quality segmented trees obtained by an automatic segmentation algorithm that were manually corrected \citep[Wytham Woods and LAUTx;][]{tockner2022automatic, calders2022laser}. The respective authors were contacted to obtain the complete unlabeled point clouds. These point clouds additionally contain non-tree points, i.e. belonging to the understory or ground, and non-annotated points, i.e. points that belong to trees but have not been annotated in the published datasets. For example, some parts of the tree crown that are hard to clearly assign to a specific tree might not have been annotated.

To obtain labels for the complete point clouds, the tree labels from the published datasets have to be propagated and the remaining points must be assigned to the classes ``non-tree'' or ``non-annotated''. This was done as follows:
\begin{enumerate}
    \item For each point in the unlabeled forest point cloud, the most common tree label within a 10~cm radius was assigned.
    \item Among the remaining unlabeled points, non-tree points were identified using proximity-based clustering: All points that were within a 30~cm distance to each other were linked and the largest connected component was labeled as non-tree points. The large grouping radius together with the high resolution of the point clouds ensured that all understory and ground points were added to the non-tree class.
    \item The points that were still unlabeled at this stage, i.e. not part of the largest connected component in the previous step, represent tree points that have not been annotated and were assigned to the non-annotated class. This information can be used to disregard these points during training. 
    \item Finally, we visually inspected the point clouds to ensure that they were adequately divided into trees, non-tree points and non-annotated points. Remaining errors were manually corrected within a feasible scope. Specifically, one large tree was not segmented in the original labeled data of \citet{calders2022laser} which we added, and the tree bases of \citet{tockner2022automatic} were corrected since they were only roughly segmented in the original labeled data.
\end{enumerate}
A visualization of the result of the above procedure is provided in Appendix~\ref{app:dataset}.
It should be noted that in several of the third party datasets the segmentation quality decreases drastically for trees smaller than 10~m. In Wytham Woods and NIBIO2 (see Table~\ref{tab:plots}) smaller trees are inconsistently labeled, i.e. sometimes as a tree and sometimes as non-tree. In LAUTx, smaller trees have severe quality limitations. For example, they are often only roughly separated from non-tree points.

\subsection{TreeLearn pipeline} \label{sec:overview}

The goal of our TreeLearn pipeline is to identify the points belonging to trees in a 3D forest point cloud and then group these points into individual trees. More precisely, given a set of 3D points $P = \{p_i = (x_i, y_i, z_i)\}_{i=1}^N$, our pipeline tackles the tasks of semantic and instance segmentation. The goal of semantic segmentation is to achieve a partition of all points into tree and non-tree points by assigning each point $p_i$ to one of the two classes. The non-tree class includes all points belonging to the forest ground and smaller vegetation in the understory. The tree class includes all points belonging to trees. What exactly constitutes a tree for the model is determined by the annotations of the datasets used for training. The goal of the instance segmentation step is to partition the tree points into $K$ (mutually exclusive) individual tree instances. The number of instances $K$ is not known a-priori and must be determined dynamically depending on the input data. The pipeline consists of the following five steps (Figure~\ref{fig:method}):
\begin{enumerate}
\item Divide the forest point cloud into smaller overlapping rectangular tiles based on the x- and y-coordinates to handle memory restrictions.
\item  Use a neural network to predict two quantities for each point in a tile: (1) an offset vector pointing towards the x- and y-coordinates of the tree base, and (2) a semantic score indicating the probability that a point belongs to a tree or not. This step is carried out separately for each tile generated in step 1. 
\item Merge the predictions for the individual tiles to obtain semantic and offset predictions for the whole input point cloud. Project the point coordinates by adding the offset.
\item Apply a clustering algorithm on these projected coordinates to partition the points into individual trees. Disregard points that are not predicted to be tree points.
\item Postprocessing: assign the remaining tree points that have not yet been assigned to a tree in step 4 by using a nearest-neighbor criterion.
\end{enumerate}

Running the complete TreeLearn pipeline on our 134~m by 125~m benchmark dataset (roughly 85 million points) took about 60~min (3 GHz CPU; 32 physical cores of which one to two are used). The runtime is primarily determined by CPU processing and can be roughly divided as following: subsampling and generating the tiles (28~min), averaging predictions (2~min), clustering (18~min), and propagation of the results to the original input point cloud (3~min). GPU calculations require roughly 9 GB of memory and runtime is comparably small (9~min). The runtime scales roughly linearly with the area of the plot. As research code, our implementation is not optimized for minimal runtime and resource usage.

\subsubsection{Step 1: Generate overlapping forest tiles} \label{sec:tiling}

The point clouds of an entire forest plot in the datasets used contain tens of millions of points. Processing a point cloud of this size at once using a neural network is impractical due to memory restrictions. To address this issue, we employ a sliding window approach to process the forest. Predictions are generated for smaller, overlapping tiles that can be processed by the neural network. This approach is inspired by \citet{ronneberger2015u} who proposed tiling to process large input images. To create the tiles, we cut out rectangular sections with a fixed edge length of 35~m along the x- and y-axis from the original forest point cloud. The edge length along the z-axis is determined by the highest point. The tiles are generated in an overlapping way with a stride of 4~m. For each point within a tile, we calculate a feature that describes the local verticality of the point cloud, which is used in later steps of the pipeline (see \secref{sec:unet} and \ref{sec:grouping}). It is calculated based on the relative magnitude of the z-component of locally computed Eigenvectors and ranges from a minimum of zero (no verticality) to a maximum of one (perfectly vertical). For a detailed description, we refer to \citet{hackel2016contour}. Prior to generating the tiles, the lidar scans are subsampled with a 10~cm voxel size, leaving only one point within a voxel. We argue that resolutions above 10~cm are not necessary for a large-scale task like tree segmentation. Subsampling also results in point clouds with the same resolution regardless of the original laser scanning resolution, provided it exceeds one point per 10~cm voxel. 

Two things should be noted: first, due to the subsampling, our algorithm does not operate on the original point cloud. However, results on the original point cloud are easily obtained by storing for each subsampled point the corresponding original points and propagating the segmentation labels at the end. This is implemented in our pipeline and has little effect on runtime. Second, the tiling procedure can also be employed if the input point cloud is not rectangular. In fact, our method can be employed on arbitrarily shaped input point clouds.

\subsubsection{Step 2: Predict semantic scores and offsets} \label{sec:unet}

\begin{figure}[t!]
   \centering
        \includegraphics[width=\textwidth]{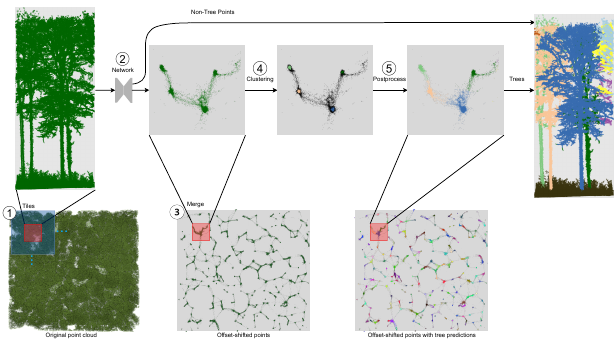}
        \vskip -0.15in
    \caption[]{The pipeline for segmenting a forest point cloud. The circled numbers correspond to the five steps of the pipeline that are described in detail in \secref{sec:overview}. Apart from the model input and output, all images depict a top-view of the forest point cloud. The blue square on the image of step~1 shows the full tile, which is used as input to the network, while predictions are generated only for the inner red area. For illustrative purposes, a side-view of only the inner part is displayed as the model input, although the network receives the entire tile. The blue dots on the image of step~1 indicate that several tiles are being processed one after the other to obtain predictions for the full input point cloud. The colored points on the image of step~4 represent the initial clustering results which only take into account points that have a sufficiently high verticality and are close to the tree base (see \secref{sec:grouping} for details). The colored points on the image of step 5 represent the final predicted trees after postprocessing.}
    \label{fig:method}
\end{figure}

After having generated smaller processable point clouds, we use a neural network to predict two quantities for every point: first, we predict the probability that a point belongs to the semantic classes tree or non-tree. As these classes are mutually exclusive, this is a binary classification problem. Second, for every point we predict an offset vector \cite{qi2019deep} that points towards the x- and y-coordinates of the respective tree base. During training, we calculate the tree base coordinate as the mean of all points between 2.75~m and 3.25~m height that have a verticality feature value (see \secref{sec:tiling}) of at least 0.6. This way, primarily trunk points are considered in the calculation of a tree base, while points from horizontal branches are ignored. Thus, the tree base can be intuitively understood as the location of the trunk at a height of three meters. The tree bases are then used to calculate the ground truth offset vectors, which are used as targets during training. During inference, the offset vectors are predicted since ground truth tree labels are not available. A height of 3~m for the tree base coordinate was chosen because, at this level, trunks are usually further apart from neighboring close trunks than at the bottom, which makes separation easier. This has already been leveraged in previous segmentation algorithms \cite{tockner2022automatic}. \fig{fig:sample_setup}a visualizes the offset vectors for a few example points.

Accurate offset prediction is possible only if the corresponding tree base is within the input tile, so the network can detect it. For trees at the outer parts of the tile, there might only be a couple of branches available. It would be impossible to perform accurate offset prediction for these points. We therefore only predict offsets for the central 8~m $\times$ 8~m portion of each tile, while the network still receives the whole \mbox{35~m $\times$ 35~m} tile as input (\fig{fig:sample_setup}b). Two things should be noted: (1) The predicted offset vector of a point within the central square can also point outside of the inner square in case that the trunk of the tree is located there. (2) Points within the central square of neighboring tiles will be projected to the same location if they belong to the same tree. This enables merging of the results (\secref{sec:merge}) and subsequent clustering of trees in a single step (\secref{sec:grouping}). Semantic predictions are also generated only for the inner square.

The size of the central square (8~m $\times$ 8~m) in relation to the complete tile (35~m $\times$ 35~m) is based on a trade-off between tree base identifiability and computational efficiency: For points within the central square, the corresponding tree base is almost always located within the tile. Only one tree in all the forest point clouds used in this work is too wide to guarantee this for its outermost points. Further decreasing the size of the central square would ensure tree base identifiability in such extreme cases, but it would also reduce computational efficiency as more tiles would need to be processed. We therefore opted against this.

Without any modifications, we use a well established neural network architecture that has been proven to be a powerful backbone for 3D scene understanding tasks \cite{vu2022softgroup, jiang2020pointgroup, schult2022mask3d}. The focus in this work is not on the network itself, but rather on its integration in a comprehensive segmentation pipeline. The network that we use for semantic and offset prediction takes as input a sparse voxel grid. First, the input space is partitioned into a 3D voxel grid where each voxel has an edge length of 10~cm. Then, for each voxel it is checked whether it contains at least one point. If it does, the voxel is active and will be considered as part of the input to the network. Only active voxels are stored. All other voxels are ignored, hence the name 'sparse voxel grid'.

\begin{figure}[t!]
\centering
\subfloat{\includegraphics[width=10cm]{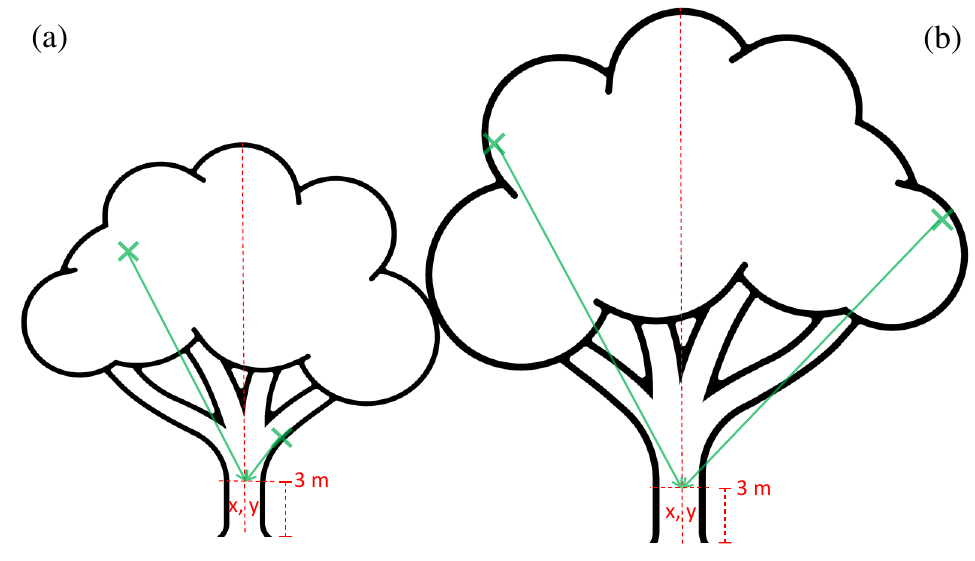}}
\hfil
\subfloat{\includegraphics[width=6cm]{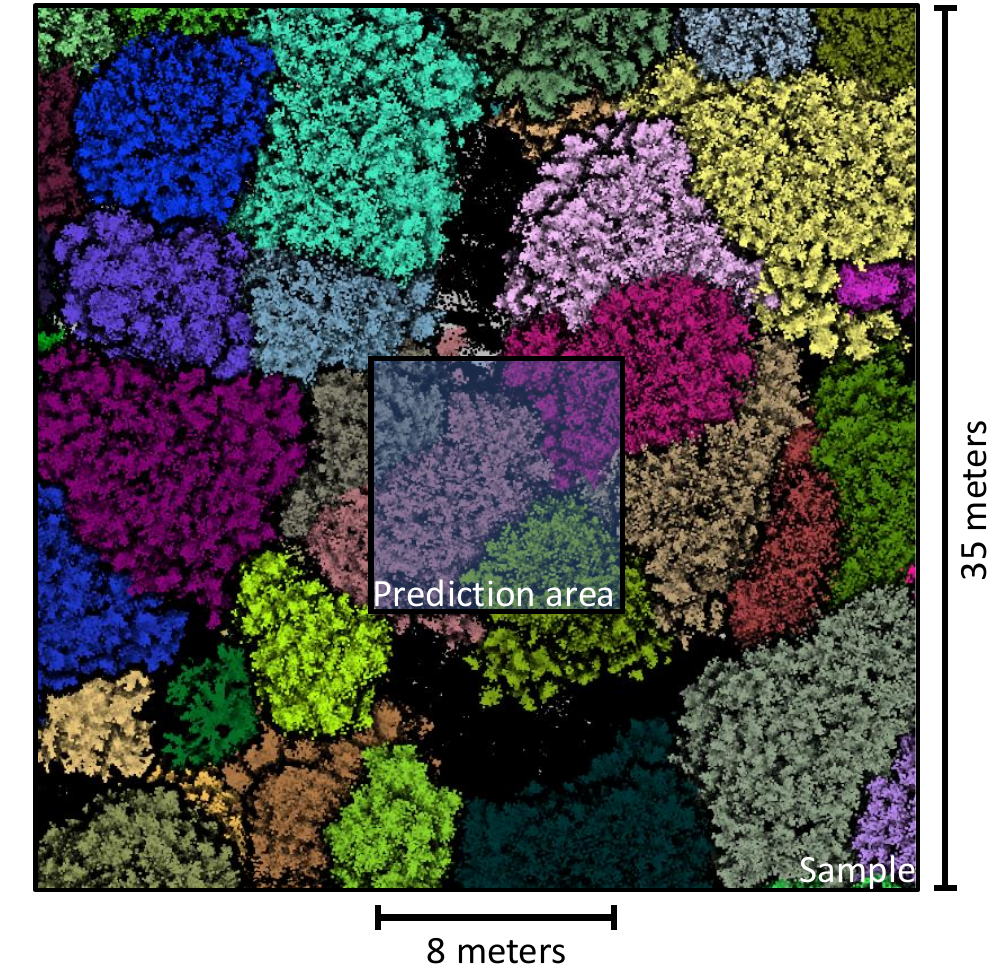}}
\caption{Visualizations regarding offset prediction. (a) depicts offsets for two example trees. The offset is the vector from a point towards the x- and y-coordinates of the tree base. The tree base is defined as the location of the trunk at a height of three meters. (b) visualizes the prediction area in relation to the whole tile. To make sure that the network has the necessary context information, predictions are only produced for the inner area of the tile. Offset prediction for points of the outer part of the tile is not possible in general since the corresponding tree bases might not be part of the tile.}
\label{fig:sample_setup}
\end{figure}

The network is a variant of the original U-Net introduced by \citet{ronneberger2015u}. The structure of a U-Net can be divided into two main steps. In the contraction step, features are computed in a hierarchical manner by gradually taking into account an increasingly larger context. In the expansion step, these features are propagated to the original points to arrive at rich voxel features that are used for semantic and offset prediction. For a more detailed explanation of the principles, we refer to the original paper \cite{ronneberger2015u}. Compared to the original U-Net, the network employed in this work follows the sparse convolution design proposed by \citet{graham20183d}. Sparse convolutions differ from regular convolutions in that an activation in the output feature channel occurs only if the voxel at the center of the convolution kernel contains a point. This keeps the number of activations constant, which avoids excessive memory consumption. Furthermore, instead of just stacking convolutions, residual blocks are used to facilitate gradient flow \cite{he2016deep}. The residual blocks consist of two 3 $\times$ 3 convolutions and a residual connection. Each contraction step of the U-Net consists of two residual blocks and a $2 \times 2$ learned pooling layer implemented by a convolution with stride two. In this work, we employ seven contraction steps. During expansion, it is ensured that the sparse structure is preserved by only propagating activations to voxels that were active during the corresponding contraction step \cite{spconv2022}.

To predict the offset and semantic class, two separate heads are used, each of which take as input point features $F$. These point features $F=\{f_1, \ldots, f_N\} \in \mathbb{R}^{N \times D}$ are obtained from the U-Net output by extracting for each of the $N$ original points the corresponding $D$-dimensional voxel feature. The semantic head maps onto semantic scores and consists of a multilayer perceptron (MLP) with two layers whose parameters are shared between points. Similarly, the offset head uses a shared three-layer MLP and maps onto offset predictions. The total number of trainable parameters of the network is roughly 30 million.

\subsubsection{Step 3: Merge tiles} \label{sec:merge}

After semantic scores and offset vectors have been predicted for the inner square of all tiles, these predictions and their corresponding points are concatenated to obtain predictions for the full forest point cloud. As the inner squares of neighboring tiles have an overlap of 4~m, multiple predictions for each point are generated. These predictions are then averaged, which reduces artifacts introduced by tiling and leads to smoother transitions between tiles. We calculate the projected coordinates $c_i = p_i + o_i$ by adding the predicted offsets ($o_i$) to the original points ($p_i$).

\subsubsection{Step 4: Identify tree instances} \label{sec:grouping}

Since the goal is to arrive at tree instances, we only consider points with a higher probability to belong to the semantic class tree in this step. If the offset prediction for these points was perfect, the projected coordinates $c_i$ would have exactly the x- and y-coordinate of the corresponding tree base. In practice, the projected points that belong to the same tree still group together, but predictions are far from perfect. Projected points are often positioned in between neighboring high density areas that represent individual trees. These points form strings connecting the high density regions (\fig{fig:method}, see image at step 3). This is the result of prediction errors for points that cannot be clearly assigned to a tree, for example if the crowns of two trees are heavily intertwined. In contrast, points belonging to the tree trunk have a much lower prediction uncertainty. Therefore, to obtain clearly separated tree clusters, we perform clustering only on points that fulfil the following two conditions: (1) the verticality feature of the point (see \secref{sec:tiling}) has a sufficiently high value of $\tau_{\textrm{vert}}$, and (2) the absolute value of the z-component of the offset prediction is at most $\tau_{\textrm{off}}$. Condition (1) already separates trunk points from other tree points to a large degree. However, some points in tree crowns also have high local verticality. Therefore, condition (2) only selects those points that are sufficiently close to the tree base, thus ensuring that crown points are excluded from clustering.

Individual tree clusters are identified by using the projected coordinates $c_i$ of the full forest point cloud that fulfil the verticality and offset condition above. These coordinates are used as input to a density-based clustering algorithm \cite{dbscan} following previous works \cite{jiang2020pointgroup, vu2022softgroup}. Specifically, an undirected graph is constructed where an edge between two points exists if their Euclidean distance is smaller than a predefined grouping radius $\tau_{\textrm{group}}$. Clusters are then obtained by identifying all connected components of the graph that consist of at least $\tau_{\textrm{min}}$ points (\fig{fig:method}, see image at step 4). A separate visualization of the clustering results is also provided in Appendix~\ref{app:clustering}. The selection of the specific values for all hyperparameters $\tau$ used for instance identification is described in \secref{sec:modelselection}.

\subsubsection{Step 5: Assign remaining points}\label{sec:remaining}

In the previous step, tree instances have been identified but a large proportion of points has not yet been assigned to individual tree instances (\fig{fig:method}, black points on image at step 4). To assign these points, we opt to use a simple strategy: the $k$ nearest neighbors of these points are determined among points that have already been assigned to trees. These neighbors are calculated based on the projected coordinates $c_i$. Therefore, the assignment of points to individual trees is done entirely based on information provided by the neural network. We set $k = 10$. Different values of $k$ lead to highly similar results.

\subsection{Training} \label{sec:training}
The U-Net described in \secref{sec:unet} was first trained using the noisy labels automatically obtained from Lidar360 (\secref{sec:trainingnoisy} below). The model obtained by training on noisy labels was then fine-tuned using the LAUTx, Wytham Woods and the For-Instance datasets (\secref{sec:trainingclean} below). Fine-tuning describes the process of using pre-trained model parameters as initialization when training with other datasets. This leads to a faster convergence compared to training from scratch. The performance of all trained models was benchmarked on the L1W and Wytham Woods datasets. 

\subsubsection{Pretraining with noisy labels} \label{sec:trainingnoisy}

To generate the training data, we cropped out random areas of size 35~m $\times$ 35~m from the 18 forest plots that have only been automatically segmented (Lidar360). These crops were then subsampled according to the procedure described in \secref{sec:tiling}. In total, \num[group-digits=integer]{20000} crops were generated.

To train the network, we calculate loss values on the predictions within the inner area of a crop, as displayed in \fig{fig:sample_setup}b. This way, the neural network almost always has enough context information. To supervise the offset prediction, we use the average L2-distance across tree points between the actual and predicted offset vector. For the semantic predictions, we use the binary cross entropy loss function and further scale it by a factor of 50 to obtain a similar magnitude to the offset loss. The offset and semantic loss are then summed to obtain the loss used for training. The U-Net was trained for \num[group-digits=integer]{150000} iterations using the AdamW optimizer \cite{adamw} with a weight decay of $10^{-3}$ and $\beta = [0.9, 0.999]$. The batch size was set to 2. We further chose a cosine learning rate schedule \cite{cosinelr} with a warm-up period of \num[group-digits=integer]{6250} iterations and a maximum/minimum learning rate of $3 \times 10^{-3}$/$5 \times 10^{-5}$. Parameters pre-trained for 3D instance segmentation from \citet{chen2021hierarchical} were used as initialization for the model. Network training was performed on a single Nvidia RTX A5000 and took roughly 40 hours.

\subsubsection{Fine-tuning} \label{sec:trainingclean}
The trained network can now already be used for segmentation. However, the performance is limited by the quality of the noisy training data. To make use of the smaller amount of manually labeled data, we perform fine-tuning of the pre-trained network with training crops generated from the LAUTx, FOR-Instance and Wytham Woods datasets.

Following the same procedure as described in \secref{sec:trainingnoisy}, \num[group-digits=integer]{30000} training crops in total were generated for the three datasets. To quantify the effect of using more training data, three fine-tuning settings were employed that differ in the datasets used (Table~\ref{tab:settings}). The \emph{small} setting uses the same 42 training plots as the conceptually similar ForAINet \cite{xiang2024automated}, allowing a direct comparison. The \emph{mid} setting was additionally trained on the validation and test data of For-Instance and the LAUTx dataset. For the \emph{large} setting, we additionally trained on the Wytham Woods plot. In all settings, the model was initialized with parameters obtained by training with noisy labels. From there, the network was trained for \num[group-digits=integer]{12500} iterations without warm-up and a maximum/minimum learning rate of $5 \times 10^{-4}$/$5 \times 10^{-5}$. Otherwise, the training specifications were the same as described in \secref{sec:training}. Fine-tuning was performed on a single Nvidia RTX A5000 and took roughly 3 hours.  A single held-out plot from NIBIO (plot 16) was used as the validation set for all settings. The model with the best offset loss on the validation set was selected as the final model for benchmarking the model performance.

\begin{table}[t!] 
\center
\caption{Fine-tuning settings. Checkmarks indicate that the respective dataset was used for fine-tuning.}\label{tab:settings}
\begin{tabular}{lccccc}
\toprule
Name & For-Instance train & For-Instance val/test & LAUTx & Wytham Woods & L1W\\
\midrule
noisy & - & - & - & - & -\\
small & \checkmark & - & - & - & -\\
mid & \checkmark & \checkmark & \checkmark & - & -\\
large & \checkmark & \checkmark & \checkmark & \checkmark & -\\
\bottomrule
\end{tabular}
\end{table}

\subsection{Hyperparameter selection} \label{sec:modelselection}

For the density based clustering step described in \secref{sec:grouping}, there are four hyperparameters. Two hyperparameters decide which points are considered in the clustering algorithm: the maximum vertical distance to the tree base was set to $\tau_{\textrm{off}}=2~\text{m}$ and the minimum required verticality feature value was set to $\tau_{\textrm{vert}}=0.6$. The verticality feature of a point can reach a maximum value of one for a perfectly vertical straight line of points. A value of $0.6$ allows for slightly slanted trunks. A value of 2~\text{m} for the maximum vertical distance from the tree base further ensures that primarily trunk points are considered. There are two additional hyperparameters that govern the behavior of the clustering algorithm: first, the minimum number of points for a valid cluster was set to $\tau_{\textrm{min}}=100$. This number was chosen based on considerations how many points a valid tree cluster might contain in extreme cases, e.g. when the tree trunk is very thin. 
Second, the grouping radius for clustering was set to $\tau_{\textrm{group}}=0.15~\text{m}$. In Appendix~\ref{app:experiment}, we analyze how the number of commission and omission errors is influenced by different grouping radii $\tau_{\textrm{group}}$.
This analysis suggests that our method can be employed out-of-the-box without extensive hyperparameter tuning.

\subsection{Evaluation metrics} \label{sec:evalmetrics}
For evaluating the quality of segmentation results, we are interested in three aspects: Separating tree points from non-tree points (semantic segmentation), detecting individual trees (instance detection) and assigning points to their corresponding tree instance (instance segmentation). As mentioned in Section~\ref{section:additional_data}, small trees are often labeled inconsistently in the existing datasets. Therefore, semantic segmentation results will be greatly influenced by which training data is used and which dataset is used for evaluation. At the same time, all of the methods compared in this work achieve a reasonable segmentation into tree and non-tree points which is why we decided to perform no evaluation of this aspect. For instance detection and instance segmentation evaluation we employ standard metrics as described in the following Sections~\ref{sec:instance_eval} and \ref{sec:instancesegmentationmetrics}. For all evaluations done in this work, we subsampled the ground truth point clouds with a voxel size of 10~cm. This is recommended since it avoids a disproportionate weighting of regions with higher point densities. For example, the trunks of ground-based scans usually have a much higher density than the tree crowns. Instructions and code for running our evaluation on arbitrary forest point clouds is publicly available in the code base of this work.

\subsubsection{Instance detection evaluation} \label{sec:instance_eval}
For instance detection evaluation we denote the set of ground truth tree instances as $\left\{ I_{\text{i}}^{\text{gt}} , i \in \{1, \dots, N_{\text{gt}} \} \right\}$ and the set of predicted tree instances as $\left\{ I_{\text{j}}^{\text{pred}} , j \in \{ 1, \dots, N_{\text{pre}} \} \right\}$. For each ground truth tree and each predicted tree, we first calculate the pointwise intersection over union (IoU):
\[
\text{IoU}(I_{\text{i}}^{\text{gt}}, I_{\text{j}}^{\text{pred}}) = \frac{TP_{\text{ij}}}{TP_{\text{ij}} + FP_{\text{ij}} + FN_{\text{ij}}}
\]
where $TP_{\text{ij}}$, $FP_{\text{ij}}$ and $FN_{\text{ij}}$ denote the number of true positive, false positive and false negative points for $I_{\text{i}}^{\text{gt}}$ and $I_{\text{j}}^{\text{pred}}$. We obtain an IoU-matrix of size $N_{\text{gt}} \times N_{\text{pre}}$. With this IoU-matrix we perform the Hungarian algorithm \cite{kuhn1955hungarian} to match ground truths with predictions. Following \citet{xiang2024automated}, we then filter out matches with an IoU smaller than 0.5 to obtain the final number of matched ground truths $N_{\text{gt}}^{\text{m}}$, unmatched ground truths $N_{\text{gt}}^{\text{u}}$ and unmatched predictions $N_{\text{pred}}^{\text{u}}$. These values are then used to define the metrics completeness, omission error, commission error, and F1-score:
\begin{align*}
    &C = \frac{N_{\text{gt}}^{\text{m}}}{N_{\text{gt}}^{\text{m}} + N_{\text{gt}}^{\text{u}}} = \frac{N_{\text{gt}}^{\text{m}}}{N_{\text{gt}}}\\
    &\errom = \frac{N_{\text{gt}}^{\text{u}}}{N_{\text{gt}}^{\text{m}} + N_{\text{gt}}^{\text{u}}} = 1 - C \\
    &\errcom = \frac{N_{\text{pred}}^{\text{u}}}{N_{\text{gt}}^{\text{m}} + N_{\text{pred}}^{\text{u}}} = \frac{N_{\text{pred}}^{\text{u}}}{N_{\text{pred}}}\\
    &F\text{1} = \frac{2 \cdot (1 - \errom) \cdot (1 - \errcom)}{(1 - \errom) + (1 - \errcom)}\\
\end{align*}

Many forest datasets have unlabeled trees at the edges. These trees are not removed during evaluation because they potentially contribute to segmentation errors, which should be taken into account. For example, part of an unlabeled tree might be falsely assigned to a prediction for a labeled ground truth tree. Apart from unlabeled trees at the edges, small trees are often not labeled. It would be inadequate to count model predictions as unmatched if they correspond to such unlabeled trees. Therefore, unmatched predictions are removed if less than half of their points belong to a labeled tree. This way, only unmatched predictions that are associated with a labeled ground truth tree are taken into account, which represent critical errors such as oversegmentation.

\subsubsection{Instance segmentation evaluation}\label{sec:instancesegmentationmetrics}
For the evaluation of instance segmentation, we want to assess how well the points of the predicted tree instances match the points of the corresponding ground truth tree instances. This is different from instance detection (Section~\ref{sec:instance_eval}), in that we are not evaluating whether a specific tree is detected at all, but how precisely each ground truth tree is segmented. We follow the evaluation protocol of \citet{xiang2024automated}. For each ground truth tree $I_{\text{i}}^{\text{gt}}$, we determine the index $max_\text{i}$ of the predicted tree with the highest IoU score:
\[
max_\text{i} = \argmax_{j=1}^{N_{\text{pre}}} \left( \text{IoU}\left( I_i^{\text{gt}} , I_j^{\text{pre}} \right) \right)
\]

This index is used to pair ground truths with predictions. It should be noted that different ground truths can be associated with the same prediction this way. For example, if a prediction is a merged tree that consists of two ground truth trees, then both of these ground truth trees will be associated with this prediction. This pairing protocol ensures that every ground truth tree can be associated with a prediction. This enables a systematic comparison of segmentation performance even if the number of matched trees according to the criterion defined in Section~\ref{sec:instance_eval} differs between methods. Following \citet{xiang2024automated}, we employ the coverage which is defined as the mean IoU across all pairs of ground truth and prediction:
\[
Cov = \frac{1}{N_{\text{gt}}} \sum_{i=1}^{N_{\text{gt}}} \text{IoU}(I_{\text{i}}^{\text{gt}}, I_{max_\text{i}}^{\text{pred}})
\]

For a more detailed performance evaluation, we also calculate the mean precision and recall:
\[
Prec = \frac{1}{N_{\text{gt}}} \sum_{i=1}^{N_{\text{gt}}} \frac{TP_{\text{i}, max_\text{i}}}{TP_{\text{i}, max_\text{i}} + FP_{\text{i}, max_\text{i}}}
\]
\[
Rec = \frac{1}{N_{\text{gt}}} \sum_{i=1}^{N_{\text{gt}}} \frac{TP_{\text{i}, max_\text{i}}}{TP_{\text{i}, max_\text{i}} + FN_{\text{i}, max_\text{i}}}
\]

Not all locations of a tree are equally difficult to segment correctly. For example, points near the trunk are usually easier to assign than points farther outside where there are many interactions with other trees. To quantify how well different parts of the trees are segmented, we partition the points of a ground truth tree and the corresponding prediction into ten subsets and calculate the metric for each subset. For each subset, the metric is then averaged across predictions. We propose two axes for partitioning: (1) horizontal distance to the trunk, and (2) vertical distance to the forest ground. For the horizontal partition, the $i$-th subset contains all points with a horizontal distance to the ground truth trunk between $\frac{i-1}{10}r$ and $\frac{i}{10}r$ where $r$ is the ground truth maximum distance. For the vertical partition, the $i$-th subset contains all points with a vertical distance to the ground between $\frac{i-1}{10}h$ and $\frac{i}{10}h$ where $h$ is the ground truth height of the tree. We set $i=1,\ldots,10$. It should be noted that points of a prediction that are farther away than $r$ and $h$ from the trunk and ground, respectively, are not taken into account in this part of the evaluation.

\section{Results and dicsussion} \label{sec:results}

TreeLearn takes unlabeled forest point clouds as input, identifies tree points and divides them into individual instances. We evaluate how accurately TreeLearn detects and segments trees on our benchmark dataset L1W and the Wytham Woods dataset, and explore whether performance improves with additional training data. We also provide a comparison with ForAINet \cite{xiang2024automated} and SegmentAnyTree \cite{wielgosz2024segmentanytree} that are conceptually similar to TreeLearn. The results of these methods were obtained via a cloud-based service (\url{https://forestsens.com/}, 26th Aug 2024). In addition, we report results of the Lidar360 algorithm \cite{Lidar360} and TLS2Trees \cite{wilkes2022tls2trees} on L1W to provide a comparison with classical algorithms based on hand-crafted features. TLS2Trees was employed with the standard hyperparameters from their work \cite{wilkes2022tls2trees}.

\subsection{Instance detection} \label{sec:instance_detection}

Regarding the instance detection results on L1W (Table~\ref{tab:testresults}), TreeLearn trained with noisy labels achieves the lowest omission and commission error rates ($\errom$=0.6\%, $\errcom$=0.6\%), closely matching the performance of Lidar360 that was used to generate the noisy labels ($\errom$=1.3\%, $\errcom$=1.3\%). Interestingly, the number of commission errors on L1W increases through fine-tuning on manually labeled data (e.g. $\errcom$=4.3\% for TreeLearn large). In Section~\ref{sec:error_analysis}, we provide a detailed error analysis for TreeLearn, where we discuss this finding. ForAINet ($\errom$=15.4\%, $\errcom$=27.1\%) and SegmentAnyTree ($\errom$=10.9\%, $\errcom$=49.1\%) exhibit the highest error rates. We conjecture that this is the result of the block-merging algorithm these methods rely on: Instance predictions are first obtained on cylinders with a radius of 8~m and subsequently merged based on an IoU-threshold. Inadequate merging could lead to over- and undersegmentation of trees (Figure~\ref{fig:instance_fig}b and c). In contrast, TreeLearn generates tree clusters for the whole input plot in a single step, leading to more accurate tree instances (Figure~\ref{fig:instance_fig}) even when trained on the same training data as ForAINet (TreeLearn small).

\begin{table}[t] 
\center
\caption{Quantitative results in~\%.} \label{tab:testresults}
\begin{tabular}{llccccccc}
\toprule
&& \multicolumn{4}{c}{Instance detection} & \multicolumn{3}{c}{Instance segmentation}\\
\cmidrule(lr){3-6}
\cmidrule(lr){7-9}
Test dataset & Method & $C$ ↑ & $\errom$ ↓ & $\errcom$ ↓ & $F\text{1}$ ↑ & $Prec$ ↑  & $Rec$ ↑  & $Cov$ ↑  \\
\midrule
\multirow{8}{*}{\centering L1W} & ForAINet & 84.6 & 15.4 & 27.1 & 78.3 & 84.1 & 86.1 & 73.8\\
& SegmentAnyTree & 89.1 & 10.9 & 49.1 & 64.8 & 90.2 & 85.2 & 77.5\\
& TLS2Trees & 96.2 & 3.8 & 4.5 & 95.8 & 87.9 & 90.5 & 79.9\\
& Lidar360 & 98.7 & 1.3 & 1.3 & 98.7 & 91.7 & 94.5 & 87.0\\
& TreeLearn noisy & 99.4 & 0.6 & \textbf{0.6} & \textbf{99.4} & 92.0 & 94.2 & 86.7\\
& TreeLearn small & 99.4 & 0.6 & 4.9 & 97.2 & 93.3 & 94.9 & 88.8\\
& TreeLearn mid & \textbf{100.0} & \textbf{0.0} & 4.3 & 97.8 & 94.8 & 96.2 & 91.3\\
& TreeLearn large & 99.4 & 0.6 & 4.3 & 97.5 & \textbf{95.1} & \textbf{96.3} & \textbf{91.8}\\
\midrule
\midrule
\multirow{4}{*}{\centering Wytham} & ForAINet & 36.8 & 63.2 & 42.9 & 44.8 & 44.5 & 84.2 & 39.7 \\
\multirow{4}{*}{\centering Woods}& SegmentAnyTree & 39.0 & 61.0 & 52.8 & 42.7 & 47.6 & 80.8 & 41.2 \\
& TreeLearn noisy & 39.9 & 60.1 & 32.2 & 50.2 & 41.2 & 71.3 & 37.9 \\
& TreeLearn small & 47.9 & 52.1 & 31.0 & 56.5 & 49.5 & 90.9 & 46.1 \\
& TreeLearn mid & \textbf{52.9} & \textbf{47.1} & \textbf{26.0} & \textbf{61.7} & \textbf{52.5} & \textbf{91.9} & \textbf{49.2} \\
\bottomrule
\vspace{0pt}
\end{tabular}
\end{table}

On the more challenging Wytham Woods dataset, performance is generally worse. TreeLearn noisy performs worst here. This is unsurprising as trees smaller than 10~m were labeled as non-tree in the training data. Wytham Woods contains many such small trees that are not detected by the model. Fine-tuning on manually labeled data that contains trees smaller than 10~m resolves this property. Nonetheless, the performance gap between TreeLearn small ($\errom$=52.1\%, $\errcom$=31.0\%) and ForAINet ($\errom$=63.2\%, $\errcom$=42.9\%) and SegmentAnyTree ($\errom$=61.0\%, $\errcom$=52.8\%) is smaller than on L1W. This is most likely due to the presence of many extremely difficult trees, for example with very closely spaced trunks (Figure~\ref{fig:instance_fig}a), where both models perform equally poorly.

\subsection{Instance segmentation}

TreeLearn achieves a superior instance segmentation performance compared to the other methods (Table~\ref{tab:testresults}). When training TreeLearn only with labels automatically generated by Lidar360 (TreeLearn noisy), our method closely matches the metrics of Lidar360. This demonstrates that the network successfully learned the segmentation rules of the algorithm. When fine-tuning TreeLearn with the same data used to train ForAINet (TreeLearn small), the coverage increases for both L1W (88.8\%) and Wytham Woods (46.1\%). On L1W, this is the case even though the number of commission errors increases, which negatively impacts the coverage. ForAINet only achieves a coverage of 73.8\% and 39.7\% on L1W and Wytham Woods, while SegmentAnyTree performs slightly better with a coverage of 77.5\% and 41.2\%. Increasing the amount of training data continuously increases the performance of TreeLearn. Compared to TreeLearn small, the coverage increases by 3\% on L1W and 3.1\% on Wytham Woods when training with all the available data. In Appendix~\ref{app:offsetloss}, we report the offset losses for all TreeLearn models.

\subsection{Detailed instance segmentation analysis}

\begin{figure}[p]
   \centering
    \includegraphics[width=0.92\textwidth]{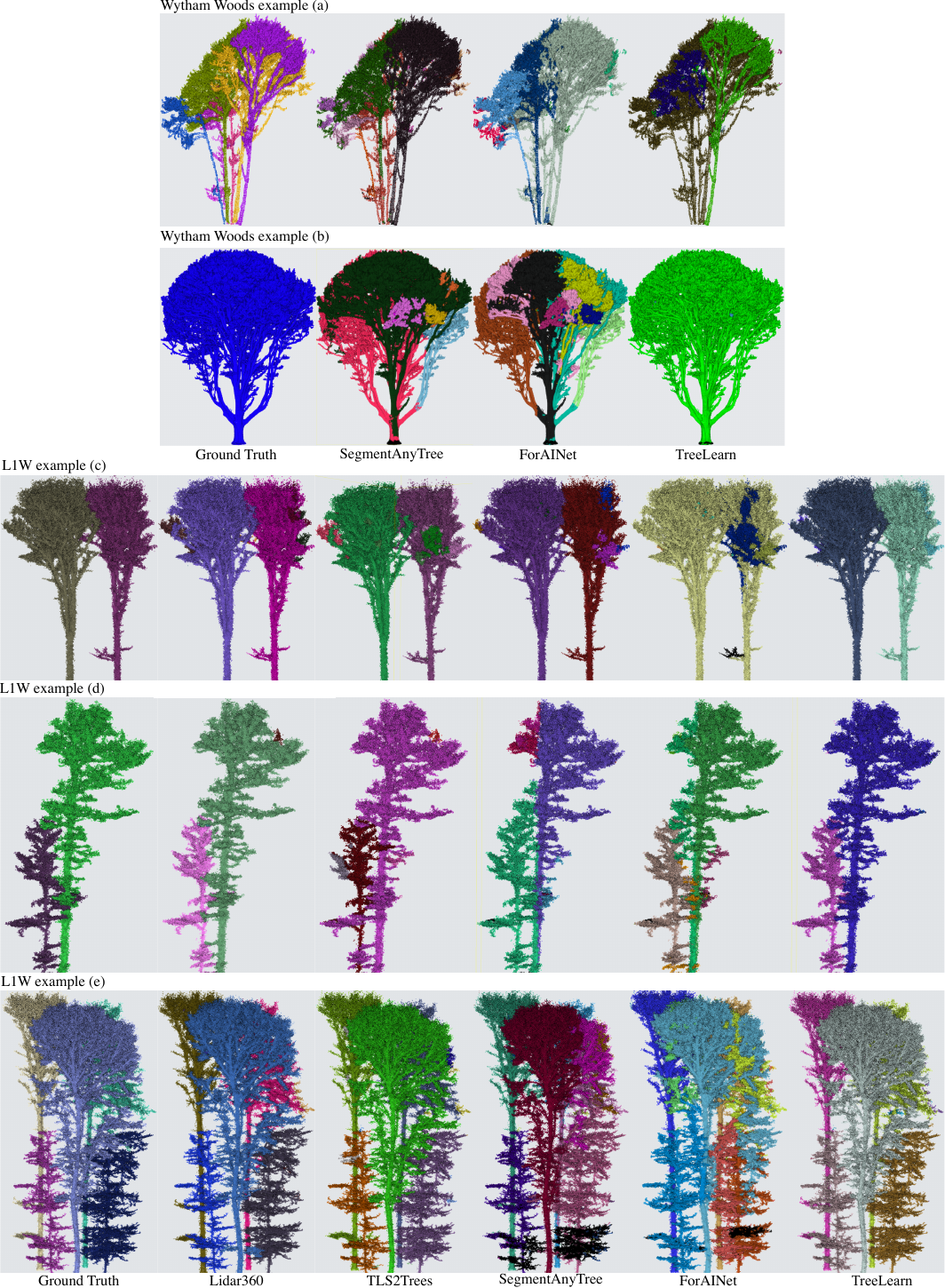}
    \caption[]{Comparison of instance segmentation results for L1W and Wytham Woods plots. TreeLearn results obtained from large (L1W) or mid setting (Wytham Woods).}
    \label{fig:instance_fig}
\end{figure}

\begin{figure}[t]
    \centering
    \includegraphics[width=\textwidth]{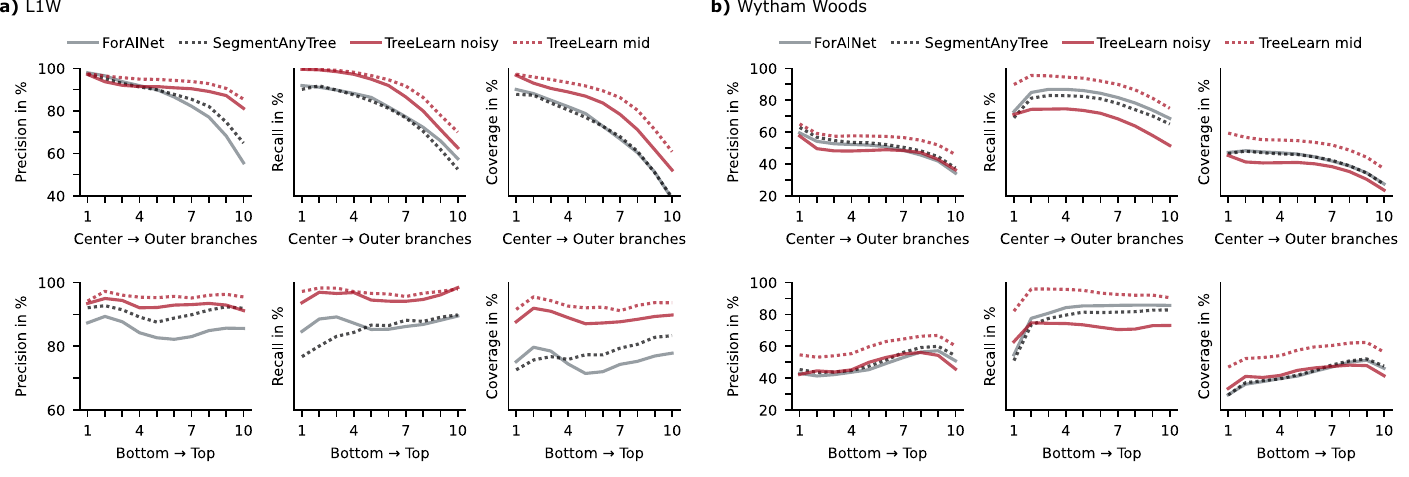} \\
    \caption{Instance segmentation performance. The top row depicts metrics for segments along the x- and y- axis, while the bottom row depicts metrics for segments along the z-axis. (a) shows results for L1W and (b) for Wytham Woods.}
    \label{fig:partitions}
\end{figure}

In this section, we take a more detailed look at the quality of the segmentations. In addition to an overall better coverage, TreeLearn achieves a better precision and recall than the comparison methods when it comes to assigning points to trees (\tableref{tab:testresults}). Irrespective of the method, recall is higher than precision. On Wytham Woods, this is caused mainly by merged trees: If a prediction is a merged tree that consists of two ground truth trees, we match both of the corresponding ground truth trees to this prediction (see Section\ref{sec:instancesegmentationmetrics}) and recall will be close to 100\%. Precision will only be around 50\% in this case, as the points of the other tree are also included in the prediction. Wytham Woods has many omission errors due to merged trees (Figure~\ref{fig:instance_fig}a). However, we also observe this pattern on L1W where merged trees are uncommon. We conjecture that, on this dataset, it is caused by the branches of larger trees protruding into smaller trees. If these branches are incorrectly assigned to the smaller tree, the precision of the smaller tree is affected considerably. On the other hand, a falsely assigned branch does not affect the recall of the larger tree to the same degree. The only exception to this pattern of results is SegmentAnyTree on L1W. This is most likely a consequence of the extraordinarily high commission error rate (49.1\%) that negatively affects recall.

A more fine-grained understanding can be achieved by evaluating the segmentation performance at different locations of a tree. For this, we partitioned the points of a ground truth tree and the corresponding prediction into subsets based on horizontal distance to the tree trunk and vertical distance to the forest ground as described in \secref{sec:instancesegmentationmetrics}. \fig{fig:partitions} shows precision, recall and coverage for the two partitions. For all methods, performance decreases for points farther away from the trunk due to increased interactions with other trees (\fig{fig:partitions}, first row). However, this trend is less pronounced for TreeLearn, indicating a certain stability with respect to the distance to the trunk. This is especially true for models fine-tuned with manually labeled data. Interestingly, the recall of ForAINet and SegmentAnyTree is already far from 100\% on both L1W and Wytham Woods for parts that are very close to the trunk, where segmentation is easiest. This can be explained by coarse segmentation mistakes, such as commission errors that make up large parts of the tree crown (e.g. \fig{fig:instance_fig}c). 

For the vertical partition on L1W (\fig{fig:partitions}a, second row), the methods tend to exhibit a visible performance low for intermediate heights, where trees tend to have their maximum width and therefore many interactions with other trees. Furthermore, the area directly above the ground poses difficulties for all methods. Potential explanations for this are low-hanging branches that are hard to clearly assign to a specific tree (also discussed in Section~\ref{sec:error_analysis}) or ambiguities between trees and understory. Wytham Woods is structurally different from L1W. It contains many smaller trees whose trunks are often very close to each other (\fig{fig:instance_fig}a). In higher parts of the forest, interactions with other trees tend to decrease since these smaller trees are not present anymore. This is reflected by the results of the vertical partition (\fig{fig:partitions}b, second row), which indicate an increasing performance from lower to higher.

\subsection{Error analysis} \label{sec:error_analysis}

\begin{figure}[t]
\centering
    {\includegraphics[width=\textwidth]{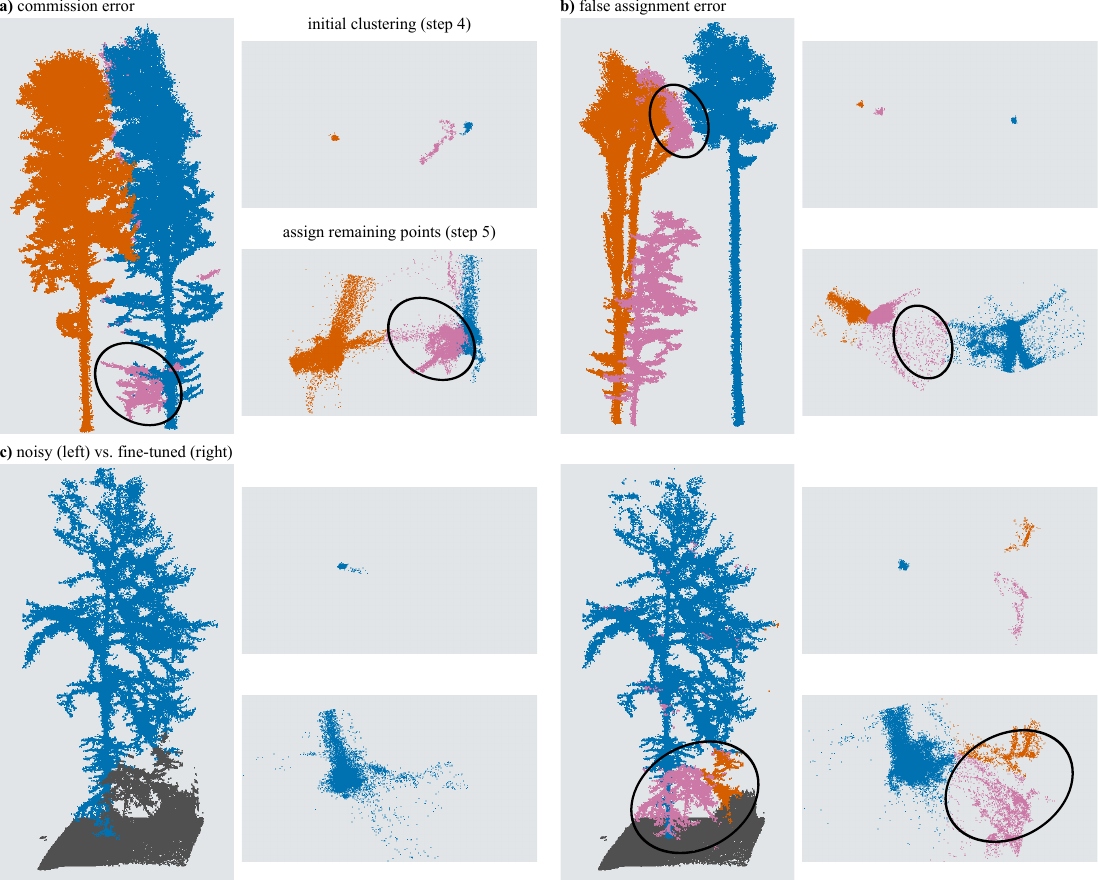}}
\caption{Failure cases in tree clustering due to systematic errors in the space of offset-shifted coordinates.}\label{fig:error}
\end{figure}

In this section, we examine failure cases of our method by taking a look at the space of offset-shifted coordinates that are used to obtain tree clusters (see Section~\ref{sec:unet}). Firstly, parts of a tree that cannot be clearly assigned to the ground truth tree base by the model can have a biased offset prediction. If enough points are projected to an incorrect location, the initial clustering algorithm (see Section~\ref{sec:grouping}) might erroneously identify these points as a tree instance. The remaining points are then assigned to this incorrect instance during postprocessing (see Section~\ref{sec:remaining}). This failure case is depicted in \fig{fig:error}a. 

Another common error arises when the model is unsure as to which tree a point belongs to. For example, this can occur for points located at the boundary of two intersecting tree crowns. In this case, these points are typically projected in-between the two tree bases resulting in a characteristic pattern of sparse connections between dense tree base clusters. If a third tree is close, uncertain points might be erroneously assigned to this tree based on spatial proximity in the space of offset-shifted coordinates. \fig{fig:error}b depicts such a failure case. Failures of this kind can lead to oddly looking mistakes where points that are far away from the ground truth tree are falsely assigned to it.

Finally, we will give an explanation for the increased number of commission errors on L1W for the fine-tuned versions of TreeLearn compared to the base version trained on noisy labels. TreeLearn noisy was trained with a dataset where trees smaller than 10~m were labeled as non-tree. In contrast, there are many smaller trees in the manually labeled datasets that have been used for fine-tuning. A qualitative inspection of the commission errors suggests that low-hanging branches might have been erroneously identified as small trees by the fine-tuned models. Points belonging to such branches were not projected to the tree base and consequently formed separate clusters (\fig{fig:error}c). This suggests that small trees are a challenge since they introduce edge cases that require additional training data for accurate prediction. 

However, simply adding more training data will not address the core challenge of this tree segmentation paradigm: Forests are complex structures, where the identification and separation of individual tree instances is a highly non-trivial task. As a consequence, neural networks trained for offset prediction will often interpolate between two trees or project points to entirely incorrect positions, complicating accurate instance prediction. To improve tree separation, one could incorporate additional parameters, such as tree height, into offset prediction. Furthermore, including some form of regularization into the offset prediction training objective could help produce more clearly separable clusters. However, solving the issues associated with this paradigm will, in our opinion, also require more general computer vision research on how to effectively discern objects with complex instance boundaries, of which trees are a prime example.

\section{Conclusion}

In this study, we introduced an automatic pipeline for segmenting individual trees from high-resolution forest point clouds, requiring no extensive hyperparameter tuning or data preprocessing in the form of ground and understory removal or terrain normalization. TreeLearn is based on a 3D U-Net for pointwise offset prediction, followed by clustering and postprocessing to derive instance predictions. Compared to two concurrent works that also employ offset prediction \cite{xiang2024automated, wielgosz2024segmentanytree}, TreeLearn avoids the need for complicated block-merging heuristics by projecting points towards their respective tree base, and clustering complete instances in a single step on the entire input point cloud. Our method achieves competitive detection and segmentation results compared to recent classical and deep-learning-based tree instance segmentation algorithms. In addition to the segmentation method, we introduced a new benchmark dataset that allows for model training and evaluation of instance segmentation results. Furthermore, we extended two previously published labeled forest point clouds \cite{tockner2022automatic, calders2022laser} to obtain a consistent input structure for neural network training. This is an important step towards the creation and comparability of powerful deep-learning-based tree segmentation algorithms. 

We highlighted several failure cases of our method that arise from difficulties in clearly separating tree instances in the space of offset-shifted coordinates. These should be tackled by future research. Furthermore, we found that using larger amounts of training data steadily increases segmentation performance. This reinforces the importance of making high-quality labeled forest datasets publicly available. A large and diverse data basis is especially important for the development of segmentation models that are applicable to point clouds with diverse laser scanning and forest characteristics. This emerging field of research \cite{wielgosz2024segmentanytree, henrich2024towards} deserves particular attention. Data-driven segmentation methods like TreeLearn will, in our opinion, play a pivotal role in this endeavour. These methods also offer the possibility to obtain additional output for detailed forest inventory. For example, \citet{xiang2024automated} demonstrated that a fine-grained semantic segmentation, e.g. into leaves and woody parts, can be directly added to tree instance segmentation models. This functionality can also be seamlessly integrated into TreeLearn without any changes to the overall pipeline. Further potential lies in the integration of tree species classification, which should be investigated by future research. All datasets that were generated in the course of this work as well as the code for training, running and evaluating TreeLearn on arbitrary forest point clouds is made publicly available.

The availability of powerful tree segmentation methods is an important prerequisite for large-scale detailed forest inventory since forest scientists and managers crucially rely on single-tree point clouds for the extraction of single-tree parameters. Such data is urgently needed to plan silviculture activities, to estimate carbon stocks or harvest yields, and for many others tasks.

\section*{Code and data availability}

The TreeLearn code is available from \url{https://github.com/ecker-lab/TreeLearn}. The automatically labeled data, our manually labeled benchmark dataset as well as trained models can be found at \url{https://doi.org/10.25625/VPMPID}. The extended versions of the LAUTx and Wytham Woods datasets are available from \url{https://doi.org/10.25625/QUTUWU}.

\section*{CRediT authorship contribution statement}
\textbf{J. H.}: conceptualization, methodology, analysis, writing - original draft, visualization, software; \textbf{J. v. D.}: conceptualization, methodology, analysis, writing - original draft, visualization, software; \textbf{D. S.}: data curation, writing - review and editing; \textbf{T. K.}: writing - review and editing; \textbf{A. E.}: supervision, writing – review and editing.

\section*{Declaration of Competing Interest}

The authors declare that they have no known competing financial interests or personal relationships that could have appeared to influence the work reported in this paper.

\section*{Declaration of Generative AI and AI-assisted technologies in the writing process}

During the preparation of this work, the authors used ChatGPT \cite{OpenAI2023} in order to paraphrase text and generate basic code for data processing and figure generation. After using ChatGPT, the authors reviewed and edited the content as needed and take full responsibility for the content of the publication.

\newpage
\bibliographystyle{unsrtnat}
\bibliography{main.bib}

\newpage
\appendix
\section[\appendixname~\thesection]{Datasets} \label{app:dataset}
Table~\ref{tab:appendix_data} contains detailed information on the 19 forest plots from our data. All of the forest plots only contain labeled trees that are at least 10~m high. Therefore, smaller trees were not taken into account in the calculation of the metrics. The tree height was calculated as the difference between the 5th-highest point and the 5th-lowest point of each tree to account for outliers. To obtain the crown diameter, tree points were first projected on the xy-plane. Then a concave hull was fitted on these coordinates. The crown diameter is defined as the farthest distance between two points on this hull. The canopy cover was calculated as the area of the hull. It should be emphasized that all plots except for L1W come with pseudo tree labels generated using the Lidar360 software. The tree parameters were extracted using these labels and should therefore only be treated as a rough approximation of the ground truth. For the L1W dataset, the tree parameters were calculated based on the 156 complete trees that were segmented manually such that they are more accurate. In L1W, the understory is sparse and there are only a few juvenile trees in the area such that the raw forest floor is mostly visible. Further information on the plots is provided by \citet{neudam2023simulation}.
\begin{table*}[h] 
\centering
\caption{Detailed information on the 19 forest plots from our data. Tree age, tree height, crown diameter and canopy cover are the mean values across all trees in a plot $\pm$ the standard deviation. \label{tab:appendix_data}}
\begin{tabularx}{\linewidth}{lYYYYYYY} 
\toprule
plot name & study area & $n$ trees & plot size ($ha$) & tree age (years) & tree height ($m$) & crown diameter ($m$) & canopy cover ($m^2$)\\
\midrule
A1N & Altstedt & 466 & 1.95 & 141 & 26.3$\pm$6.9 & 10.5$\pm$3.6 & 54.9$\pm$36.5 \\
A1W & Altstedt & 511 & 1.42 & 67 & 22.0$\pm$5.5 & 8.8$\pm$2.7 & 39.5$\pm$23.1 \\
O1N & Oppershofen & 160 & 1.22 & 163 & 39.2$\pm$9.4 & 13.4$\pm$4.5 & 88.3$\pm$53.2 \\
O1W & Oppershofen & 266 & 1.46 & 124 & 27.5$\pm$10.0 & 10.7$\pm$4.0 & 61.5$\pm$38.6 \\
G1N & Göttingen & 320 & 1.07 & 143 & 30.1$\pm$7.3 & 9.5$\pm$3.1 & 45.4$\pm$26.2 \\
G1W & Göttingen & 450 & 2.15 & 124 & 27.8$\pm$6.5 & 10.9$\pm$2.7 & 61.2$\pm$26.6 \\
G2N & Göttingen & 496 & 1.56 & 162 & 23.9$\pm$7.5 & 10.0$\pm$4.0 & 53.5$\pm$39.3 \\
G2W & Göttingen & 265 & 1.00 & 149 & 22.9$\pm$10.2 & 9.4$\pm$3.9 & 49.7$\pm$33.9 \\
G3N & Göttingen & 355 & 1.56 & 137 & 29.6$\pm$8.4 & 10.8$\pm$3.3 & 58.2$\pm$31.0 \\
G3W & Göttingen & 332 & 1.48 & 129 & 27.1$\pm$8.0 & 10.8$\pm$3.1 & 59.4$\pm$29.1 \\
G4N & Göttingen & 318 & 1.08 & 133 & 27.2$\pm$9.1 & 10.3$\pm$3.5 & 54.1$\pm$33.9 \\
G4W & Göttingen & 585 & 2.19 & 130 & 29.2$\pm$8.6 & 10.7$\pm$3.5 & 57.6$\pm$36.2 \\
L1N & Lübeck & 453 & 1.69 & 151 & 26.3$\pm$8.8 & 11.1$\pm$3.7 & 65.2$\pm$41.1 \\
L2N & Lübeck & 233 & 1.29 & 135 & 30.8$\pm$8.9 & 12.7$\pm$4.0 & 81.8$\pm$43.4 \\
L2W & Lübeck & 401 & 1.67 & 92 & 25.0$\pm$8.3 & 10.6$\pm$3.4 & 57.8$\pm$33.9 \\
LG1 & Lübeck & 407 & 1.14 & 128 & 33.9$\pm$7.6 & 8.8$\pm$2.7 & 38.0$\pm$21.5 \\
LG2 & Lübeck & 332 & 1.53 & 127 & 28.3$\pm$9.0 & 12.2$\pm$3.9 & 74.0$\pm$39.0 \\
LG3 & Lübeck & 315 & 1.59 & 149 & 25.5$\pm$5.7 & 11.9$\pm$3.4 & 71.0$\pm$35.8 \\
L1W & Lübeck & 156 & 1.16 & 136 & 30.0$\pm$6.5 & 12.6$\pm$3.0 & 85.2$\pm$36.8 \\
\bottomrule
\end{tabularx}
\end{table*}

\begin{figure}[h]
\centering
    {\includegraphics[width=\textwidth]{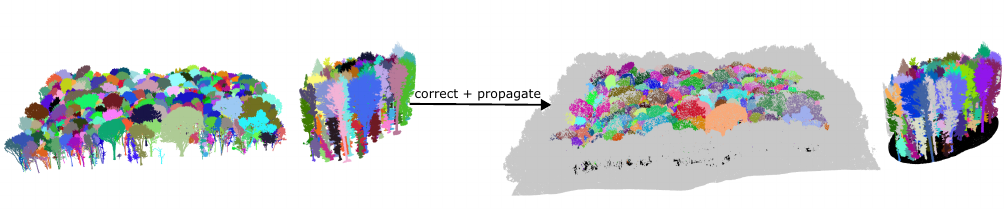}}
\caption{Propagation of segmentation labels to the complete forest point clouds for Wytham Woods (left) and LAUTx (right). Non-tree and unlabeled points were added. Furthermore minor mistakes were corrected as can be seen for the tree bases in LAUTx.}\label{fig:propagation}
\end{figure}

\section[\appendixname~\thesection]{Clustering} \label{app:clustering}
\begin{figure}[H]
\centering
    {\includegraphics[width=\textwidth]{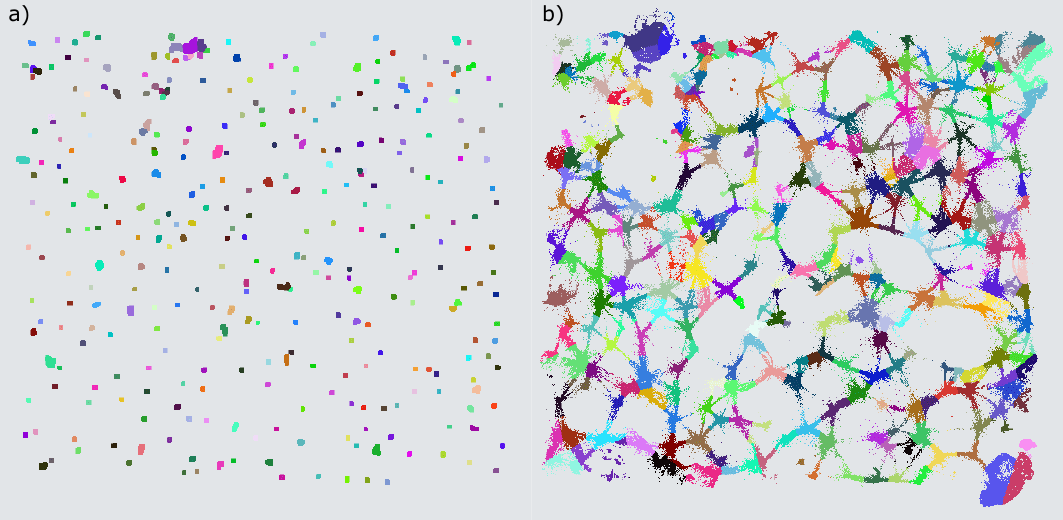}}
\caption{Clustering of the offset-shifted coordinates on the L1W dataset. (a) shows the initial clustering result that only takes into account points that are close to the tree base and have a sufficiently high verticality (see \secref{sec:grouping} for details) and (b) shows the results after assigning the remaining points (see \secref{sec:remaining} for details).}\label{fig:cluster_space}
\end{figure}

\section[\appendixname~\thesection]{Robustness analysis for grouping radius} \label{app:experiment}

\begin{figure}[h]
\centering
    {\includegraphics[width=\textwidth]{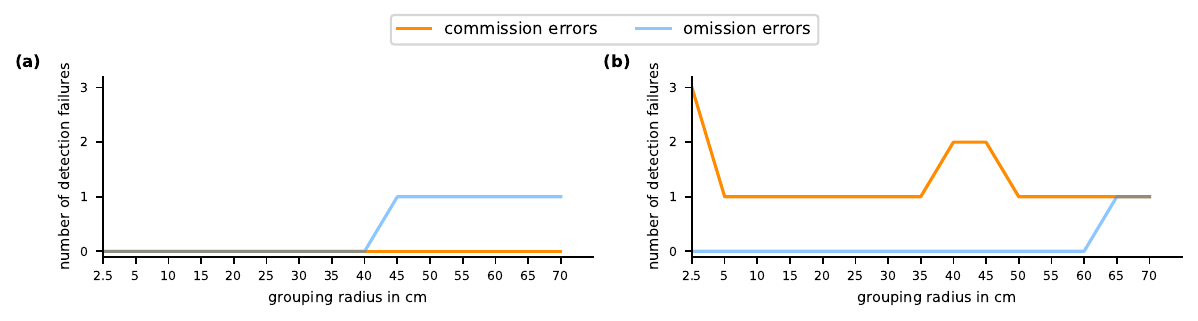}}
\caption{Number of commission and omission errors depending on the grouping radius for (a) TreeLearn trained with Lidar360 labels and (b) TreeLearn that was additionally fine-tuned.}\label{fig:experiment}
\end{figure}

The density-based clustering algorithm described in \secref{sec:grouping} is controlled by the two hyperparameters $\tau_{\textrm{min}}$ and $\tau_{\textrm{group}}$. The selected minimum number of points for a valid cluster $\tau_{\textrm{min}}=100$ is relatively low. Therefore, there is the risk of incurring commission errors if a tree prediction does not actually constitute a tree but only a fragment. To reduce the number of commission errors, the grouping radius can be increased. However, at some point this might lead to omission errors due to tree clusters merging together. To quantify this trade-off between commission and omission errors, we conducted an experiment where we vary the grouping radius $\tau_{\textrm{group}}$ between 2.5~cm and 70~cm and measure the corresponding detection errors on L1W. For the fine-tuned model, the expected tendency for clusters to break apart could be observed for a very small grouping radius of 2.5~cm, while this was not the case for the model trained on noisy labels (\fig{fig:experiment}, orange line). This can be attributed to the fact that, in the fine-tuned setting, a single tree-cluster is more likely to be split into distinguishable sub-clusters due to a more precise offset prediction. Furthermore, the expected tendency for clusters to merge could be observed for both settings (\fig{fig:experiment}, blue line). However, the number of detection failures is in general very low and not strongly influenced by changes in the grouping radius, which suggests that our method is robust to changes in this hyperparameter. For practical purposes, we recommend setting the grouping radius relatively low (somewhere in the range of 0.1~m to 0.2~m) since commission errors can be easily dealt with by simply merging fragments together, while merged trees cannot be that easily recovered. It should be noted that this experiment was conducted with an older version of the model and a slightly different evaluation protocol, which is why the reported error rates differ from what is reported in Section~\ref{sec:results}.

\section[\appendixname~\thesection]{Offset loss} \label{app:offsetloss}

\begin{table}[h] 
\center
\caption{Offset loss of TreeLearn models in m. Value in brackets shows the loss when disregarding the z-component.\label{tab:offsetloss}}
\begin{tabular}{llc}
\toprule
Test dataset & Method & Offset Loss ↓ \\
\midrule
\multirow{4}{*}{\centering L1W} 
& TreeLearn noisy & 2.41 (2.27) \\
& TreeLearn small & 2.72 (2.30)\\
& TreeLearn mid & 2.47 (2.27)\\
& TreeLearn large & 2.40 (2.25)\\
\midrule
\multirow{2}{*}{\centering Wytham Woods} 
& TreeLearn noisy & 6.15 (5.60) \\
& TreeLearn small & 6.35 (5.84)\\
& TreeLearn mid & 6.20 (5.80)\\
\bottomrule
\vspace{0pt}
\end{tabular}
\end{table}

Table~\ref{tab:offsetloss} shows the offset loss for the TreeLearn models trained with different amounts of training data. The value of the offset loss when only considering the x- and y-component of the offset vector is also shown since the z-component is ignored during clustering. For the fine-tuned models, a lower offset loss corresponds to a better performance, as is expected. TreeLearn noisy has the lowest offset loss for both L1W and Wytham. However, it should be noted that this model was trained with an older version of our pipeline where the input point cloud was additionally filtered using statistical outlier removal. Therefore, the loss is computed on a different set of points and the better results cannot be clearly interpreted. In general, the offset loss is better on L1W, reflecting the better segmentation performance on this dataset. The validation loss during fine-tuning on plot 16 of the NIBIO dataset was between 0.5~m and 0.7~m for all models. This value is much lower which might be explained by the similarity of the training data to it. Most point clouds used during training have also been captured via UAV. It might also be a consequence of this plot being relatively easy to segment.

\end{document}